\documentclass{article}

\usepackage{arxiv}

\usepackage[utf8]{inputenc}
\usepackage{amsmath, amssymb, amsfonts, bm}
\usepackage{cite}
\usepackage{booktabs}
\usepackage{hyperref}
\usepackage{multirow}
\usepackage{graphicx}
\usepackage{multicol}
\usepackage{subcaption}
\usepackage{caption}

\title{Spectral Manifold Regularization for Stable and Modular Routing in Deep MoE Architectures}
\author{Ibrahim Delibasoglu}
\date{January 2026}

\begin{document}

\maketitle

\begin{abstract}
Mixture of Experts (MoE) architectures provide a powerful paradigm for scaling neural networks, yet they are frequently hindered by "expert collapse," where a subset of experts dominates the routing manifold, leading to reduced modularity and significant catastrophic interference during adaptation. In this paper, we propose the Spectrally-Regularized Mixture of Experts (SR-MoE), a novel framework designed to enforce structural modularity through geometric constraints on the routing manifold. By introducing a dual-objective penalty—constraining the spectral norm to bound Lipschitz constants and regularizing the stable rank to maintain high-dimensional feature diversity—we ensure that routing decisions remain stable and "surgical."

We evaluate our approach across two architectural scales and varying dataset complexities using a modular one-shot adaptation task. Our results demonstrate that while traditional linear gating fails as network depth increases—experiencing accuracy drops of up to 4.72\% due to expert entanglement—the SR-MoE maintains structural integrity with a mean interference of only -0.32\%. Furthermore, we show that our spectral constraints facilitate positive knowledge transfer, allowing for localized expert updates without global performance decay. This framework provides a general-purpose solution for developing high-capacity, modular neural networks capable of stable, lifelong learning across diverse domains. 
\end{abstract}

\section{Introduction}

Modern deep neural networks often suffer from catastrophic interference and the \textit{loss of plasticity} during one-shot/few-show learning tasks. Loss of plasticity is a phenomenon where a model's ability to adapt to new information diminishes over time as weight matrices deviate from their beneficial initialization properties \cite{lewandowski2024learning}. When a traditional dense model is updated with a single new sample, the resulting gradient flow typically affects the entire parameter space, potentially degrading performance on previously learned distributions while simultaneously "hardening" the network against future updates.

To address this, we propose a modular architecture that employs a learnable, spectrally-constrained clustering stage to route data to specialized sub-networks, or ``experts.'' The core of our approach is the use of a \textbf{Spectral-Regularized Router}. Unlike standard gating mechanisms that may suffer from expert collapse or routing instability, our network is constrained by spectral norm and stable rank regularization at gates managing the input distribution of experts. This ensures that the mapping from input space to cluster assignments is Lipschitz-continuous and robust to noise.

By maintaining the singular values of the routing weights near unity, we preserve the \textit{gradient diversity} necessary for continual trainability \cite{lewandowski2024learning}. This enables a surgical approach to one-shot learning: the stable router correctly identifies the "path" for a new sample, allowing us to update only the weights of the relevant specialized expert. This localized update preserves the integrity of the remaining global model while ensuring that the network remains "plastic" and ready for subsequent novel tasks.

Unlike existing spectral regularization methods that seek to globally stabilize deep stacks, we propose Targeted Spectral Anchoring of the routing manifold. By penalizing deviations from a target spectral norm and stable rank, we ensure that the routing stage maintains a high-rank latent representation that is both sensitive to class-specific features and robust to the local perturbations of one-shot updates."

\section{Related Works}

\subsection{Regularization and Spectral Analysis in Deep Learning}
Regularization is central to enhancing the generalization of deep neural networks. Classical approaches include $\ell_2$ (weight decay) and $\ell_1$ regularization, while Dropout \cite{srivastava2014dropout} prevents feature co-adaptation. Label smoothing \cite{muller2019does} improves generalization by preventing overconfident predictions, and data-level methods like Mixup \cite{zhang2023mixupexplainer} and CutMix \cite{yun2019cutmix} synthesize training examples through interpolation. Beyond these, spectral normalization \cite{miyato2018spectral} stabilizes training by bounding the network's Lipschitz constant, while orthogonality-based methods such as Parseval networks \cite{cisse2017parseval} and orthogonal regularization \cite{bansal2018can, huang2017orthogonal} improve gradient flow and reduce redundancy.

The spectral properties of neural networks have been shown to correlate strongly with generalization. Yoshida and Miyato \cite{yoshida2017spectral} demonstrated that constraining the singular value spectrum enforces function-space smoothness. Spectral penalty methods \cite{bartlett2017spectrally} and efficient spectral analysis for convolutions \cite{sedghi2018singular} further enable practical stability in vision models. This connects to Martin and Mahoney's \textit{Heavy-Tailed Self-Regularization} (HT-SR) theory \cite{mahoney2019traditional}, where the power-law exponent $\alpha$ of singular value distributions correlates with generalization: lower $\alpha$ values ($\alpha \approx 2$) signal strong implicit regularization, while deviations indicate over- or under-fitting. Our work extends this spectral perspective to the routing mechanism in mixture-of-experts, using spectral constraints to enforce stable cluster boundaries in the assignment space.

\subsection{Spectral Methods for Clustering and Representation Learning}
Spectral clustering provides a principled approach to partitioning data based on the eigenvectors of graph Laplacians. \textbf{SpectralNet} \cite{shaham2018spectralnet} scales this approach by using neural networks to learn embeddings that approximate these eigenvectors. Recent advances like \textbf{Double-stage Feature-level Clustering} \cite{badjie2025double} demonstrate that pre-clustering features before expert assignment significantly reduces noise impact. In this work we reconceptualize the routing problem: rather than applying spectral clustering as a separate stage, we embed spectral properties directly into the routing network through spectral norm regularization. This encourages the router to learn representations with natural cluster structure, making expert assignment more stable and geometrically meaningful.

\subsection{Mixture of Experts and Routing Stability}
Modular routing was established by Jacobs et al. \cite{jacobs1991adaptive} via gating networks. Modern large-scale implementations like \textbf{Sparsely-Gated MoE} \cite{shazeer2017outrageously} use noisy top-k gating to scale model capacity efficiently. However, these approaches often ignore the geometric stability of the routing latent space. The inherent volatility of dynamic routing—where small input variations cause disproportionate assignment changes—hinders consistent expert specialization.

\textbf{StableMoE} \cite{dai2022stablemoe} directly addresses this instability through a two-stage strategy involving router distillation to reduce routing volatility, ultimately freezing the router to create static data paths. While effective, this approach sacrifices routing adaptability. Our work offers a complementary solution: instead of fixing the router after distillation, we use spectral regularization to achieve stability \textit{during} joint training of router and experts. This maintains plasticity while ensuring geometric smoothness in the assignment function, preventing experts from needing to "chase" changing assignments and enabling better handling of novel one-shot data.

\subsection{Modularity for Few-Shot and Continual Learning}
Modular architectures show promise for adaptation to new tasks with minimal data. Our work builds upon \textbf{Prototypical Networks} \cite{snell2017prototypical}, replacing static prototypes with active, spectrally-regularized experts that can adapt to new classes. This is conceptually related to \textbf{Sub-Network Routing (SNR)} \cite{ma2019snr} for preventing negative transfer in multi-task learning. Crucially, by applying principles of spectral regularization to sustain trainability \cite{lewandowski2024learning}, we ensure that localized updates to an expert for a new task do not corrupt the global feature space or routing policy. This provides a path toward lifelong learning without catastrophic forgetting, as the spectrally-constrained router maintains stable boundaries between expert regions even as experts themselves adapt.

\section{Methodology}

We propose a modular deep learning framework centered on a Spectrally-Regularized Mixture of Experts (SR-MoE). While we demonstrate its efficacy through image classification and one-shot adaptation tasks, the core contribution of this work is a generalized architectural improvement to the MoE routing mechanism.

\begin{figure}
    \centering
    \includegraphics[width=1.0\linewidth]{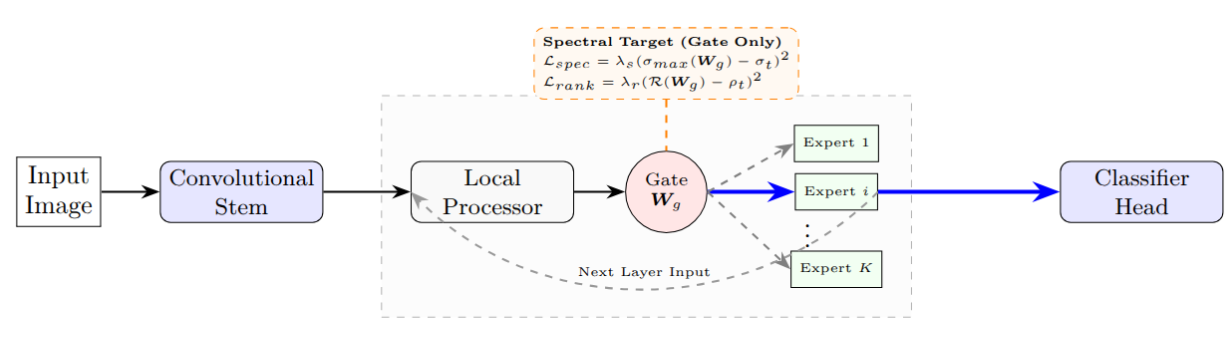}
    \caption{Deep SR-MoE Architecture. The model processes inputs through $N$ successive layers. In each layer, a bank of $K$ experts is available. The \textit{Spectral Regularization} is strictly applied to the routing weights $\bm{W}_g$ in every layer to ensure manifold diversity. Surgical updates are performed by backpropagating only through the active expert chain (blue path).}
    \label{fig:overview}
\end{figure}

Standard Mixture of Experts models often suffer from "Expert Collapse," where the gating network converges to a narrow subspace, effectively under-utilizing the model's total capacity and leading to catastrophic interference during fine-tuning. Our framework addresses this by regularizing the routing manifold's geometry, ensuring that the gating mechanism remains stable, diverse, and responsive to novel distributions. By anchoring the router's weights via spectral constraints, the approach enforces structural modularity that is agnostic to the specific downstream task, making it applicable to any domain requiring high-plasticity adaptation or modular feature partitioning. As illustrated in Figure \ref{fig:overview}, the system utilizes a feature-extraction backbone followed by specialized SR-MoE stages that partition the latent space through prototype-based clustering.

\subsection{Architectural Design}

The model is composed of three primary components: a convolutional stem, a sequence of $N$ stacked MoE layers, and a global classification head.

\subsubsection{Convolutional Feature Extraction}

The input image $\bm{x} \in \mathbb{R}^{C \times H \times W}$ is first processed by a convolutional stem. This stage consists of sequential layers of strided convolutions, non-linear activations, and max-pooling to extract high-level spatial features. An adaptive average pooling layer followed by a linear projection maps these features into a latent embedding $\bm{z}_0 \in \mathbb{R}^d$, which serves as the input to the MoE stages.

\subsubsection{MoE Layer Mechanics}

Each MoE stage $l \in \{1, \dots, N\}$ consists of a local processor, a prototype-based router, and a set of $K$ parallel experts.

\begin{enumerate}\item \textbf{Local Processor:} Before routing, the latent vector is refined through a transformation $\phi(\cdot)$ comprising a linear layer, ReLU activation, and Layer Normalization. This prepares the manifold representation for the gating decision: $\bm{z}'_l = \phi(\bm{z}_{l-1})$.\item \textbf{Expert Execution:} The layer output is computed as a weighted sum of expert transformations. Each expert $E_i$ is a multi-layer perceptron (MLP) that processes the refined latent vector:\begin{equation} \bm{z}l = \sum{i=1}^K w_i E_i(\bm{z}'_l) \end{equation}\end{enumerate}

\subsubsection{Prototype-based Clustering and Routing}

Instead of traditional dot-product gating, we employ a distance-based routing mechanism that interprets gating as a geometric clustering task. Each MoE layer maintains a set of learnable prototypes $\{\bm{\mu}_1, \dots, \bm{\mu}_K\}$ in the latent space. The router computes the Euclidean distance between the refined latent embedding $\bm{z}'$ and each prototype. The routing weights $\bm{w}$ are derived using a softmax over the negative distances:\begin{equation} w_i = \frac{\exp(-|\bm{z}' - \bm{\mu}_i|2 / \tau)}{\sum{j=1}^K \exp(-|\bm{z}' - \bm{\mu}_j|_2 / \tau)} \end{equation}where $\tau$ is a temperature hyperparameter controlling the sharpness of the routing decision. This formulation ensures that inputs are routed to experts based on their proximity to specific manifold clusters.

\subsection{Spectral Manifold Regularization}

To ensure the routing stage maintains a robust and plastic latent space, we introduce a composite spectral penalty. This approach anchors the routing manifold by constraining both the energy and the dimensionality of the gating weights, preventing the "rank collapse" often observed in deep mixture-of-experts networks. We specifically target the linear gate parameters $\bm{W}_l$ at each layer $l$.

\subsubsection{Spectral Norm Penalty}
We bound the Lipschitz constant of the routing decision by penalizing the deviation of the weight matrix's largest singular value $\sigma_{max}$ from a predefined target energy $\sigma_{t}$. For a weight matrix $\bm{W}_l$, the penalty is defined as:

\begin{equation}
\mathcal{L}_{spec\_norm}(\bm{W}_l) = \left( \sigma_{max}(\bm{W}_l) - \sigma_{t} \right)^2
\end{equation}

where $\sigma_{max}(\bm{W}_l) = \|\bm{W}_l\|_2$ is the spectral norm. This ensures that the router remains in a sensitive gradient regime, preventing numerical instability during rapid one-shot adaptation.

\subsubsection{Stable Rank Regularization}

To prevent the router from collapsing its decision space onto a single dominant feature, we regularize the \textit{stable rank} $\mathcal{R}(\bm{W}_l)$, which serves as a robust proxy for the numerical rank. It is defined as the ratio of the squared Frobenius norm to the squared spectral norm:

\begin{equation}\mathcal{R}(\bm{W}_l) = \frac{|\bm{W}_l|F^2}{\sigma_{max}(\bm{W}_l)^2}\end{equation}
We enforce a target feature diversity $\rho_t$ using a squared error penalty:

\begin{equation}
\mathcal{L}_{rank}(\bm{W}_l) =  \left( \mathcal{R}(\bm{W}_l) - \rho_t \right)^2
\end{equation}

This ensures the gating layer utilizes a high-dimensional subspace, allowing the prototypes to remain distinct and well-separated in the latent manifold.

\subsubsection{Expert Diversity (Load Balancing)}

To ensure global expert utilization and prevent "lazy routing," we also use an additional load-balancing loss $\mathcal{L}_{div}$ based on the coefficient of variation ($CV^2$) of the expert importance. Let $P_i$ be the average importance of expert $i$ across a batch of $B$ samples: $P_i = \frac{1}{B} \sum_{b=1}^{B} w_{i}^{(b)}$. The diversity loss is defined as:\begin{equation}\mathcal{L}_{div} = \left( \frac{\text{std}(\bm{P})}{\text{mean}(\bm{P})} \right)^2\end{equation}This term penalizes non-uniform expert selection, forcing the gates/routers to distribute its capacity across the available experts.

\subsection{Total Multi-Objective Loss}

The final objective function used for training and one-shot updates integrates the task-specific cross-entropy $\mathcal{L}_{task}$ with our structural constraints:

\begin{equation}
\mathcal{L}{total} = \mathcal{L}{task} + \alpha \sum_{l=1}^{N} \left[ \mathcal{L}_{spec\_norm}(\bm{W}_l) + \mathcal{L}{rank}(\bm{W}_l) \right] + \beta \mathcal{L}_{div}
\end{equation}where $\alpha$ is the spectral scaling factor and $\beta$ is the diversity scale. By optimizing this joint objective, we preserve the structural integrity of the experts, enabling the model to perform surgical path updates without inducing catastrophic interference.

\begin{figure}[ht]
     \centering
     \begin{subfigure}[b]{0.33\textwidth}
         \centering
         \includegraphics[width=\linewidth]{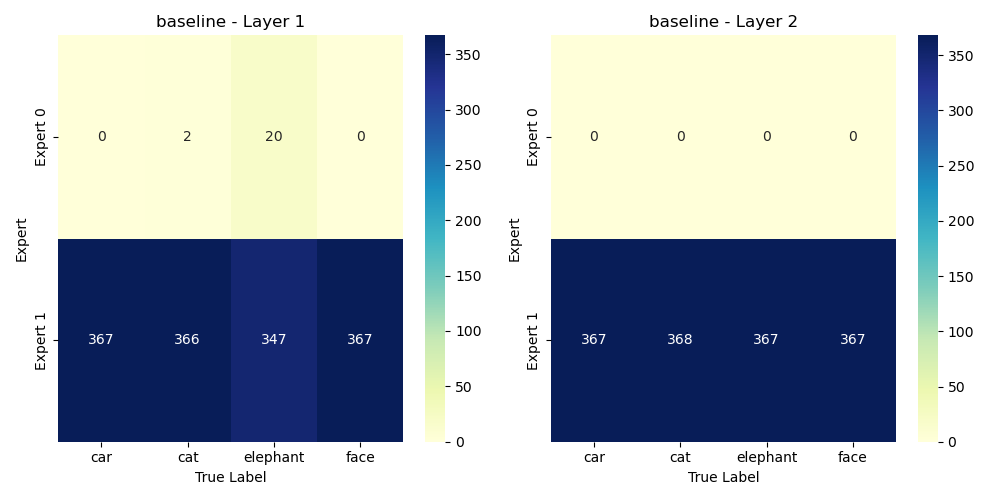}
         \caption{Baseline}
         \label{fig:base525}
     \end{subfigure}
     \hfill
     \begin{subfigure}[b]{0.33\textwidth}
         \centering
         \includegraphics[width=\linewidth]{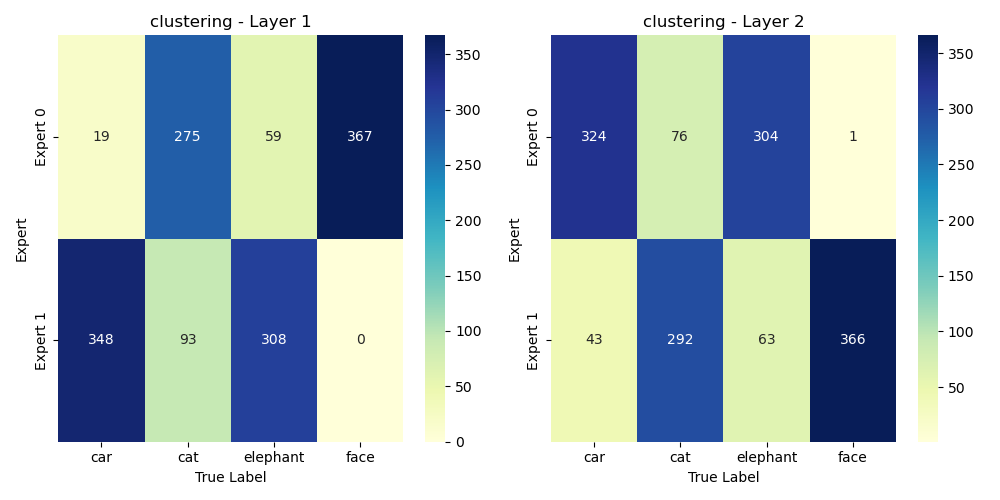}
         \caption{Clustering}
         \label{fig:clus525}
     \end{subfigure}
     \hfill
     \begin{subfigure}[b]{0.33\textwidth}
         \centering
         \includegraphics[width=\linewidth]{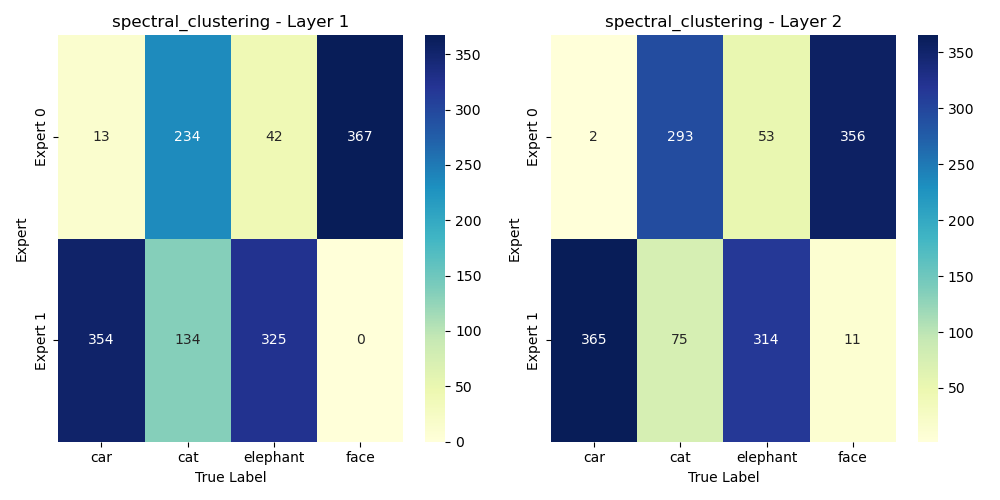}
         \caption{Spectral clustering (ours)}
         \label{fig:spec525}
     \end{subfigure}
     
     \vspace{0.2cm}
     \caption{Expert distribution on the small dataset (N=525). The baseline shows complete path collapse (all data routed through a single expert). Clustering improves load balancing, and spectral clustering begins to separate semantic concepts into distinct expert pathways as detailed in Section \ref{sec:one_shot_interference}.}
     \label{fig:dataset525}
\end{figure}

\begin{figure}[ht]
     \centering
     \begin{subfigure}[b]{0.32\textwidth}
         \centering
         \includegraphics[width=\linewidth]{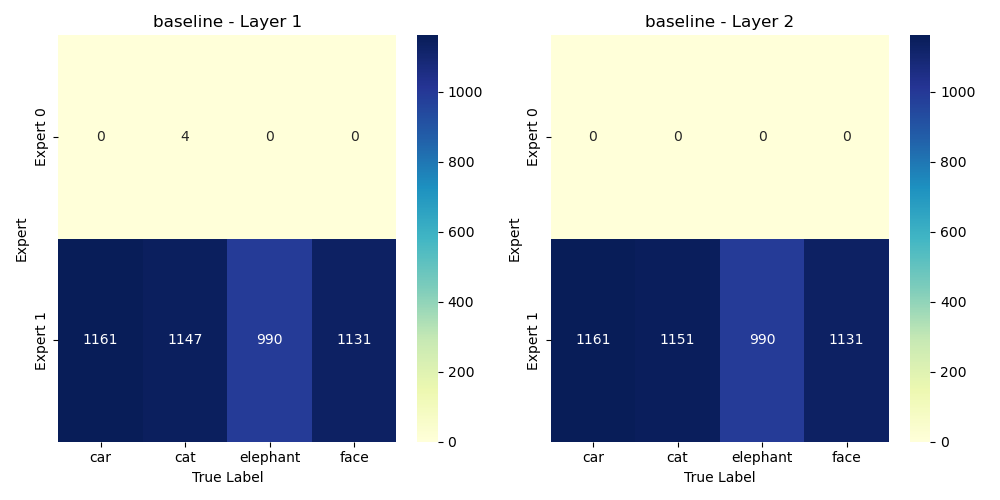}
         \caption{Baseline}
         \label{fig:base1600}
     \end{subfigure}
     \hfill
     \begin{subfigure}[b]{0.32\textwidth}
         \centering
         \includegraphics[width=\linewidth]{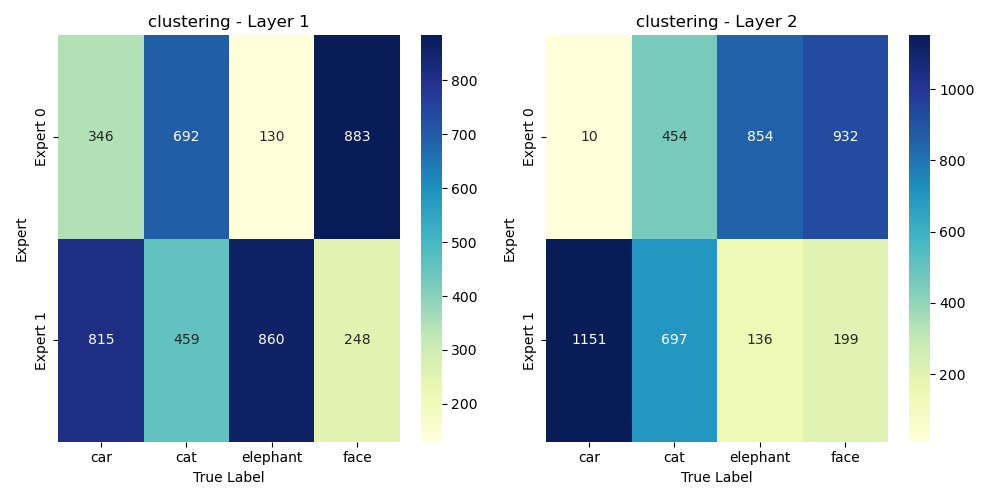}
         \caption{Clustering}
         \label{fig:clus1600}
     \end{subfigure}
     \hfill
     \begin{subfigure}[b]{0.32\textwidth}
         \centering
         \includegraphics[width=\linewidth]{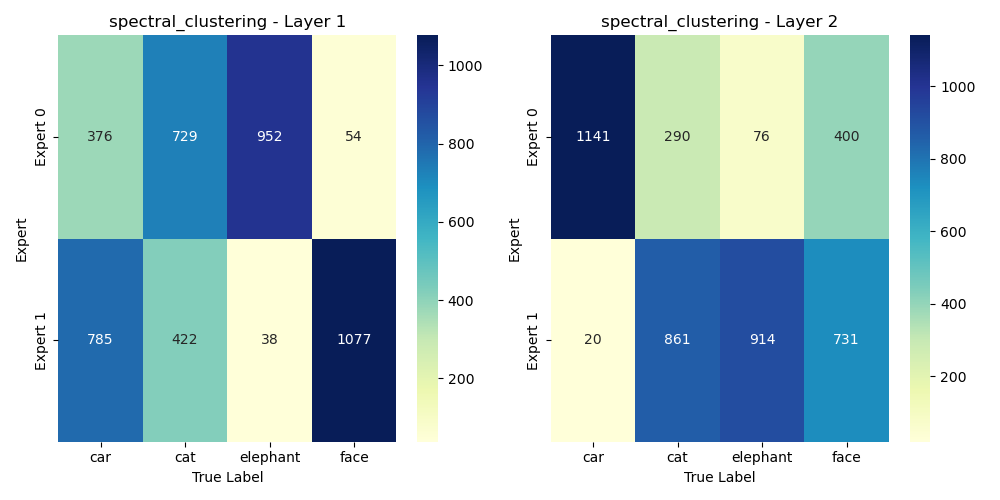}
         \caption{Spectral clustering (ours)}
         \label{fig:spec1600}
     \end{subfigure}
     
     \vspace{0.2cm}
     \caption{Path utilization on the large dataset (N$\approx$ 1600). With sufficient data, the spectral model successfully utilizes the full architectural capacity ($2\times2$ experts), mapping each semantic class to a distinct expert circuit. Clustering shows improved load balancing over baseline, but lacks the structured specialization of spectral routing as detailed in Section \ref{sec:one_shot_interference}.}
     \label{fig:dataset1600}
\end{figure}

\section{Experimental Evaluation}

\subsection{Experimental Setup and Data Scaling}
We initially trained a classification model on four distinct categories: \textit{Car, Cat, Elephant,} and \textit{Face}. The experiment was conducted in two phases to evaluate scalability: 
\begin{enumerate}
    \item \textbf{Small-Scale Phase:} Each class contained 525 samples (split 70/15/15).
    \item \textbf{Relatively Large-Scale Phase:} Dataset was expanded to approximately 1600 samples per class to stabilize manifold formation.
\end{enumerate}

For the modular one-shot adaptation test, we selected 25 novel images per class that were excluded from the original training set. To prevent the "single-sample collapse" common in one-shot learning, we utilized an \textbf{Anchor-Batch} strategy: the model was updated using a  novel sample and a small memory batch from the original training set. All reported results represent the average accuracy across the test data set following these modular updates.

\subsection{Path Utilization Analysis}
We visualize the expert selection distribution to assess structural modularity. A "surgical" model should ideally demonstrate category-specific path clustering.

\begin{figure}[ht]
    \centering
    
    \begin{subfigure}[b]{0.95\textwidth}
        \centering
        \includegraphics[width=0.95\linewidth]{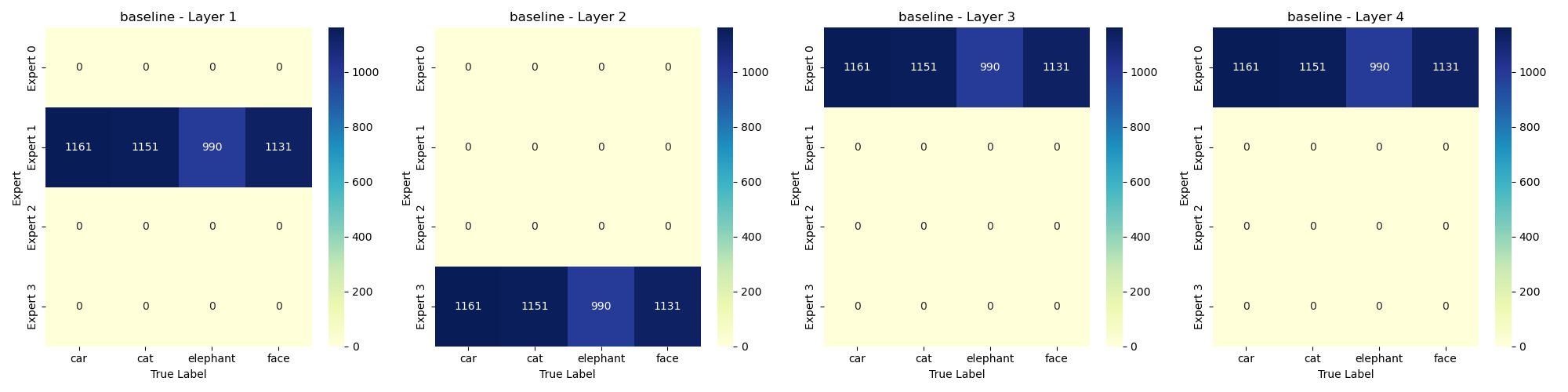}
        \caption{Baseline}
        \label{fig:base1600}
    \end{subfigure}
    
    \vspace{0.3cm}
    
    \begin{subfigure}[b]{0.95\textwidth}
        \centering
        \includegraphics[width=0.95\linewidth]{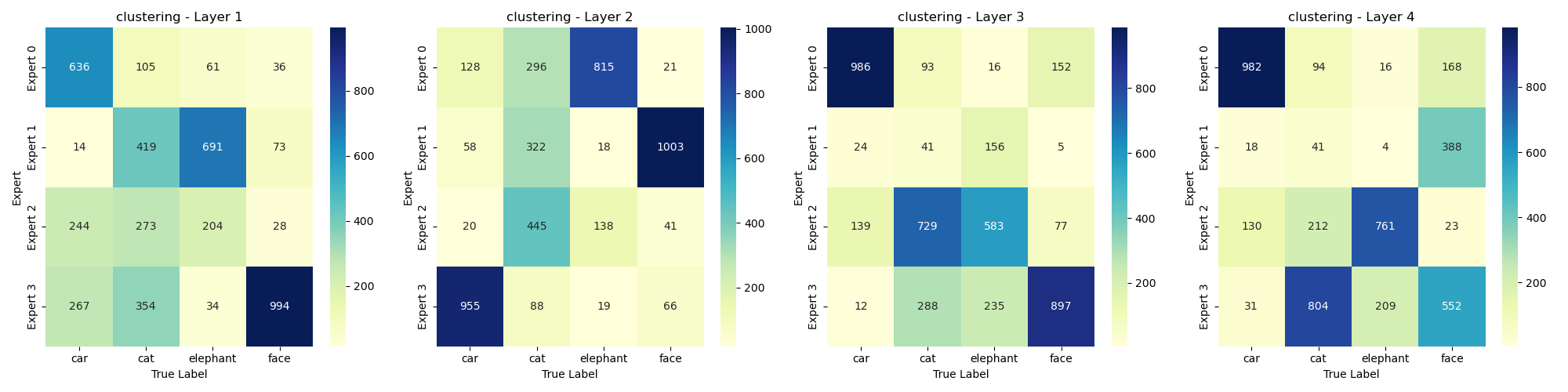}
        \caption{Clustering}
        \label{fig:clus1600}
    \end{subfigure}
    
    \vspace{0.3cm}
    
    \begin{subfigure}[b]{0.95\textwidth}
        \centering
        \includegraphics[width=0.95\linewidth]{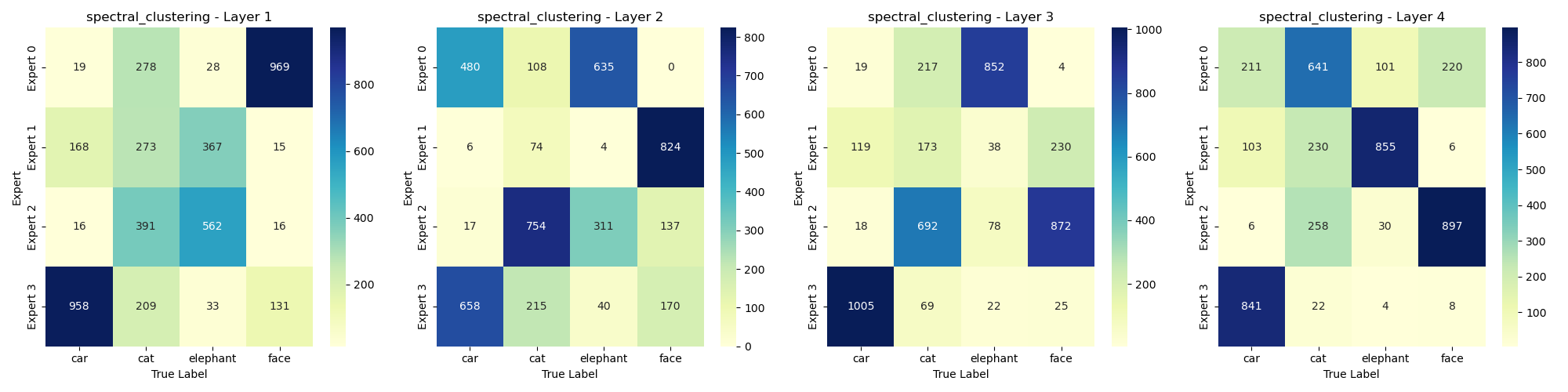}
        \caption{Spectral Clustering (ours)}
        \label{fig:spec1600}
    \end{subfigure}
    
    \vspace{0.2cm}
    \caption{Expert utilization patterns across routing methods (4×4 experts). \textbf{(a)} Baseline shows poor load balancing. \textbf{(b)} Clustering improves distribution but lacks structured specialization. \textbf{(c)} Spectral clustering provides both balanced load and semantic specialization, mitigating interference as analyzed in Section~\ref{sec:one_shot_interference}.}
    \label{fig:dataset1600}
\end{figure}

The transition from 525 to 1600 samples reveals a critical threshold for manifold stability. In the initial phase ($N=525$), all models exhibited a degree of path overlap, with the Baseline utilizing only a single path for all classes (Figure \ref{fig:dataset525}a). However, upon scaling to $N=1600$, the Spectral Clustering router achieved full architectural expression. As shown in Figure \ref{fig:dataset1600}c, each of the four categories migrated to a non-overlapping path in the $4 \times 4$ expert grid. This structural separation directly correlates with the one-shot performance: while the Baseline suffered a -8.39\% accuracy drop due to weight overwriting, the Spectral model maintained a near-zero interference delta (-0.21\%), as the one-shot updates were confined to experts that remained dormant for other classes as detailed in section \ref{sec:one_shot_interference}.

\begin{figure}
    \centering
    \includegraphics[width=1.0\linewidth]{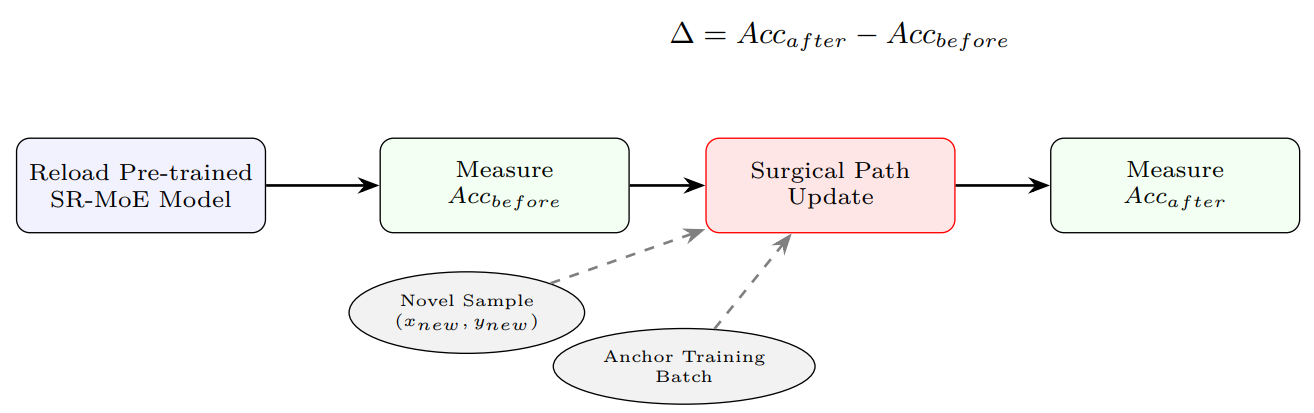}
    \caption{One-Shot Experimental Workflow. The model undergoes a surgical update using a single novel sample anchored by a training batch. The resulting Accuracy Delta ($\Delta$) quantifies the degree of catastrophic interference.}
\label{fig:one_shot_workflow}
\end{figure}

\subsection{One-Shot Interference Analysis}
\label{sec:one_shot_interference}

To evaluate the structural integrity of the learned experts, we perform a modular one-shot adaptation test. As illustrated in Figure \ref{fig:one_shot_workflow}, the process begins by establishing a baseline test accuracy on a fresh model. We then select a novel image $x_{new}$ and perform a ``surgical'' weight update.To maintain the global distribution and prevent the manifold from collapsing onto a single point—a risk when updating with a single sample—we utilize an \textbf{Anchor-Batch update strategy}. In this approach, the gradient is computed using the novel sample concatenated with a small auxiliary batch from the original training set. Finally, we re-evaluate the model on the entire test set to measure the accuracy delta ($\Delta$), which serves as our primary metric for quantifying catastrophic interference.

\subsubsection{Expert Collapse in Baseline Architectures}

The experimental results demonstrate a critical trade-off between initial generalist performance and long-term architectural stability. As shown in Tables \ref{tab:large_dataset} and \ref{tab:large_dataset_layers4}, the Baseline model exhibits a consistent phenomenon of \textit{Path Collapse}. By routing 100\% of all samples through a single static expert chain, the model maximizes initial accuracy on the base dataset (84.23\% in the shallow configuration). However, this lack of structural diversity creates extreme vulnerability during one-shot adaptation. Since all category features are "entangled" within the same weights, fine-tuning the model for a single new sample (e.g., a \textit{Face}) inadvertently overwrites the features required for other classes. This leads to significant catastrophic forgetting, with the Baseline experiencing a mean interference of -1.41\% in shallow networks, which escalates to a devastating -4.72\% in deep configurations.

\subsubsection{Surgical Plasticity via Spectral Regularization}In contrast, our proposed \textit{Spectral Clustering} approach enforces a high-rank routing manifold that effectively partitions the network into category-specific circuits. While the initial accuracy in the shallow model (82.97\%) is slightly lower than the Baseline—a result of experts becoming specialists rather than generalist ensembles—the modular benefits become apparent during one-shot training.By isolating updates to the "winning path," our model achieves a positive mean $\Delta$ (+0.41\%) in the 2-layer test, indicating that the model can learn new information without degrading existing knowledge. In the case of the \textit{Car} category, we observe \textbf{Positive Transfer} (+1.17\%), where surgical adaptation actually improves global test performance.

\subsubsection{Scalability to Deep MoE Architectures}The true efficacy of Spectral Regularization is revealed in the 4-layer, 4-expert configuration. As task complexity and model depth increase, the Baseline model's performance collapses under the weight of interference, losing 8.39\% accuracy on the \textit{Face} category. Conversely, Spectral Clustering emerges as the superior architecture, achieving both the highest pre-update accuracy (80.44\%) and the highest stability (mean $\Delta$ of -0.32\%).This demonstrates that for deep Mixture-of-Experts systems, spectral constraints are not merely optional regularizers but essential mechanisms for maintaining \textit{Structural Plasticity}. By anchoring the routing logic in a stable, high-dimensional manifold, our system ensures that deep networks remain modular and capable of one-shot adaptation without global structural decay.

\begin{table}[h]
\centering
\caption{MoE performance with one-shot training (2 Layers, 2 Experts, and N $\approx$ 1600 per class)}
\label{tab:large_dataset}
\begin{tabular}{@{}lccc@{}}
\toprule
Metric & Baseline & Clustering & \textbf{Spectral (Ours)} \\ \midrule
Avg. Initial Acc & \textbf{84.23\%} & 83.28\% & 82.97\% \\
\midrule
\textit{Accuracy Delta ($\Delta$)} & & & \\
--- Car          & -1.15\% & +0.89\% & \textbf{+1.17\%} \\
--- Cat          & -0.98\% & \textbf{+0.42\%} & +0.36\% \\
--- Elephant     & -1.61\% & \textbf{+1.01\%} & +0.29\% \\
--- Face         & -1.91\% & -0.46\% & \textbf{-0.20\%} \\
\midrule
\textbf{Mean Delta} & -1.41\% & \textbf{+0.47\%} & +0.41\% \\
\textbf{Path Diversity} & 1 Path (Collapsed) & 4 Paths & 4 Paths \\
\bottomrule
\end{tabular}
\end{table}

\begin{table}[ht]
\centering
\caption{Deep MoE Performance with one-shot training (4 Layers, 4 Experts, and N $\approx$ 1600 per class).}
\label{tab:large_dataset_layers4}
\begin{tabular}{@{}lrrr@{}}
\toprule
\textbf{Evaluation Metric} & \textbf{Baseline} & \textbf{Clustering} & \textbf{Spectral (Ours)} \\ \midrule
Pre-Update Base Accuracy   & 71.61\%           & 76.76\%             & \textbf{80.44\%}         \\ \midrule
\textit{One-Shot Accuracy Delta ($\Delta$)} & & & \\
--- Car       & -2.34\%     & \textbf{-0.03\%}    & -0.31\%    \\
--- Cat          & -2.38\%    & -1.40\%   & \textbf{-1.01\%}   \\
--- Elephant    & -5.75\%     & -1.54\%    & \textbf{+0.26\%}  \\
--- Face       & -8.39\%     & -1.91\%    & \textbf{-0.21\%}  \\ \midrule
\textbf{Mean Interference} & -4.72\%     & -1.22\%    & \textbf{-0.32\%}  \\
\textbf{Path Utilization}  & Static (Collapse) & Stochastic  & \textbf{Modular}  \\ \bottomrule
\end{tabular}
\end{table}

\begin{figure*}[ht]
    \centering
    \begin{subfigure}[b]{0.32\textwidth}
        \centering
        \includegraphics[width=\textwidth]{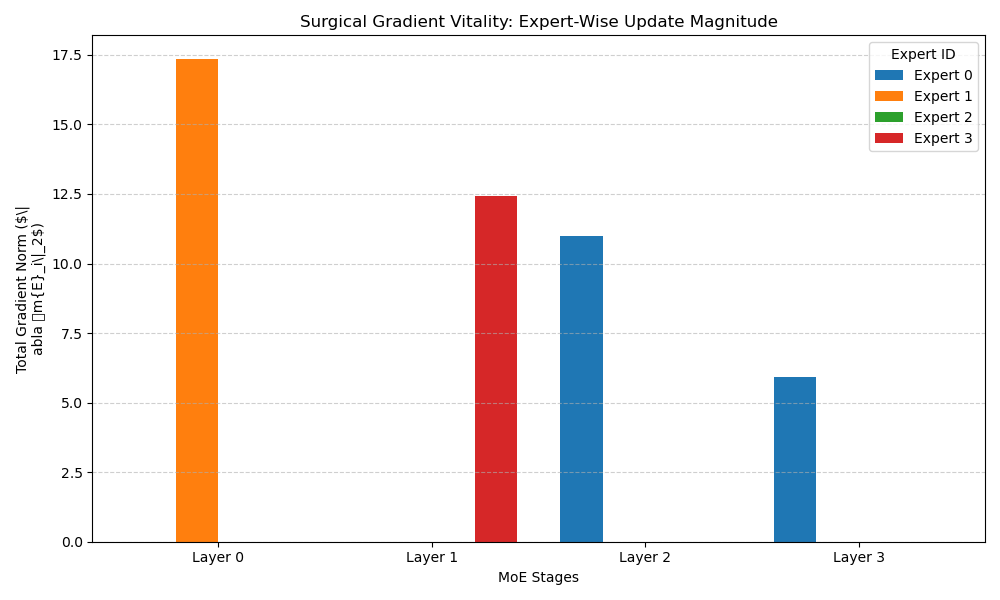}
        \caption{Baseline (Collapsed)}
        \label{fig:grad_base}
    \end{subfigure}
    \hfill
    \begin{subfigure}[b]{0.32\textwidth}
        \centering
        \includegraphics[width=\textwidth]{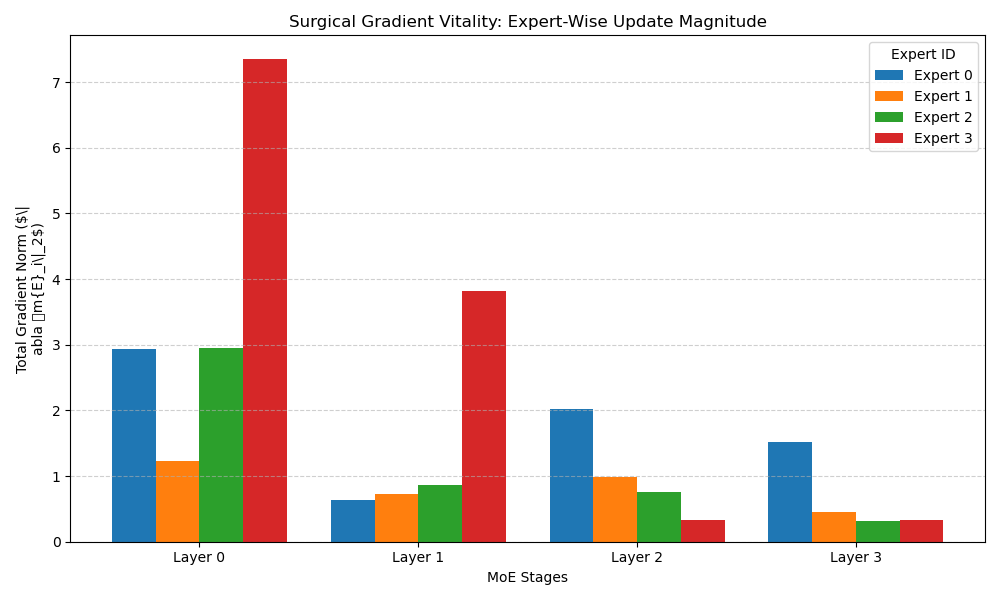}
        \caption{Distance Clustering}
        \label{fig:grad_clus}
    \end{subfigure}
    \hfill
    \begin{subfigure}[b]{0.32\textwidth}
        \centering
        \includegraphics[width=\textwidth]{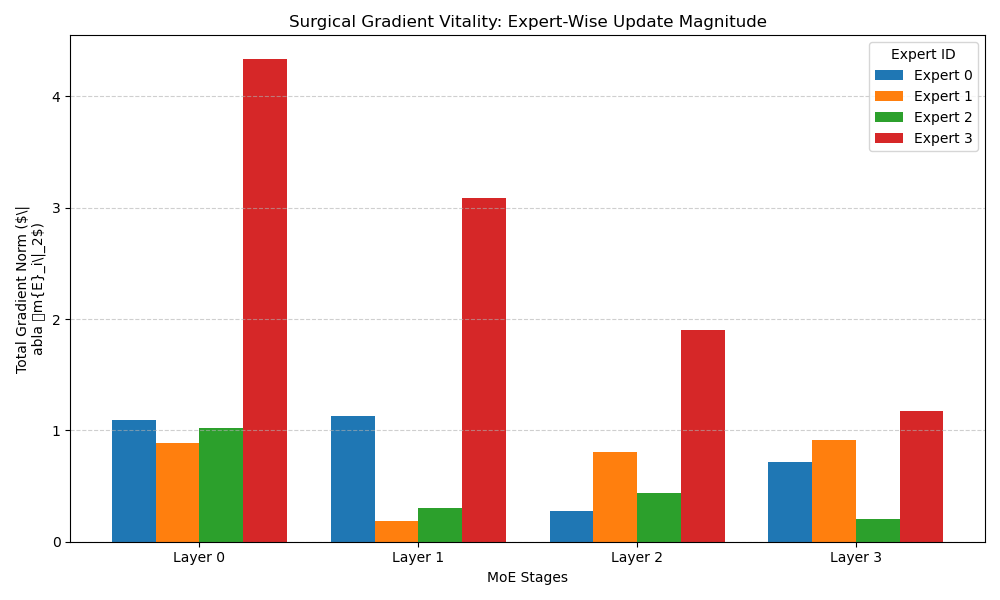}
        \caption{\textbf{Spectral Clustering (Ours)}}
        \label{fig:grad_spec}
    \end{subfigure}
    \caption{Gradient Vitality Analysis across 4 MoE layers. The magnitude of the gradient norm per expert reveals the "surgical" nature of the updates. Baseline exhibits extreme path sparsity (collapse), while Spectral Clustering maintains a balanced and modular gradient flow.}
    \label{fig:gradient_vitality_comparison}
\end{figure*}

\textbf{Gradient Vitality and Path Sparsity:} To empirically validate the modular behavior of our framework, we measured the gradient norm magnitude ($\|\nabla \bm{E}_i\|_2$) for each expert during a single one-shot update. As shown in Figure \ref{fig:gradient_vitality_comparison}, the Baseline model exhibits extreme gradient sparsity, where the updates are confined to a single "surviving" expert per layer (e.g., Expert 1 in Layer 0 and Expert 3 in Layer 1), with remaining experts receiving negligible gradients ($< 10^{-11}$). This confirms the existence of path collapse. Gradient norm is also so high compared to the others. In contrast, the Spectral Clustering approach demonstrates a structured distribution of gradient vitality. While updates are still "surgical" in the sense that they follow a specific routing path, the gradient energy is distributed across a more diverse set of experts. Specifically, in Layer 0 of the Spectral model, the gradient is prioritized toward Expert 4 (magnitude 4.33), yet the remaining experts remain "warm" and accessible. This indicates that Spectral Regularization prevents the weights from becoming numerically dead, ensuring that the model retains the capacity to learn diverse features without the binary "on/off" failure state observed in the baseline.

\section{Conclusion}

In this work, we presented a Spectrally-Regularized Mixture of Experts (SR-MoE) framework designed to bridge the gap between high-capacity neural networks and modular structural plasticity. Drawing inspiration from the biological brain's ability to segregate information into specialized functional regions, our approach utilizes spectral norm and stable rank constraints to enforce a diverse and non-collapsed routing manifold.

Our comparative analysis reveals that while distance-based clustering can mitigate total expert collapse, it often lacks the structural stability required for deep architectures, leading to stochastic path selection and interference. In contrast, our spectral approach anchors the routing manifold, providing a significantly clearer and more robust partitioning of the latent space. Ultimately, this research provides a powerful strategy for building modular neural networks. By ensuring that expert isolation is not just achieved but mathematically preserved, we pave the way for scalable, lifelong learning systems that can adapt to new knowledge with the same localized efficiency seen in the human brain.

\bibliographystyle{unsrt}
\bibliography{references}

\end{document}